\pdfoutput=1

\documentclass[11pt]{article}

\usepackage[]{ACL2023}

\usepackage{times}
\usepackage{latexsym}

\usepackage[T1]{fontenc}

\usepackage[utf8]{inputenc}

\usepackage{microtype}

\usepackage{inconsolata}
\usepackage{multicol}
\usepackage{booktabs}
\usepackage{colortbl}
\usepackage[fleqn]{amsmath}
\usepackage{multirow}
\usepackage[normalem]{ulem}
\useunder{\uline}{\ul}{}
\usepackage{hyperref}
\usepackage{graphicx}
\usepackage{subfigure}
\usepackage{fixltx2e}
\usepackage{xcolor}
\usepackage{algorithm}
\usepackage{algorithmic}
\usepackage{setspace}
\usepackage{amsfonts}
\usepackage[most]{tcolorbox}

\newcommand{\ie}{\textit{i.e.}}
\newcommand{\eg}{\textit{e.g.}}
\newcommand{\ours}{\textsc{ConvAug}}
\newcommand{\todo}[1]{\{\textcolor{blue}{\textbf{TODO}}\}}

\title{Generalizing Conversational Dense Retrieval via LLM-Cognition \\ Data Augmentation}

\author{Haonan Chen$^{1}$, Zhicheng Dou$^{1}$\thanks{$^{*}$Corresponding author.}, Kelong Mao$^1$ \\ \textbf{Jiongnan Liu$^1$, Ziliang Zhao$^1$} \\
        $^1$Gaoling School of Artificial Intelligence, Renmin University of China \\ 
        \texttt{\{hnchen,dou\}@ruc.edu.cn} \\
}

\newtcolorbox[list inside=prompt]{prompt}[1][]{
    colbacktitle=black!60,
    coltitle=white,
    fontupper=\footnotesize,
    boxsep=5pt,
    left=0pt,
    right=-1pt,
    top=0pt,
    bottom=0pt,
    boxrule=1pt,
    #1,
}
\begin{document}
\maketitle

\begin{abstract}

Conversational search utilizes muli-turn natural language contexts to retrieve relevant passages.
Existing conversational dense retrieval models mostly view a conversation as a fixed sequence of questions and responses, overlooking the severe data sparsity problem -- that is, users can perform a conversation in various ways.
Consequently, they often struggle to generalize to diverse conversations in real-world scenarios.
In this work, we propose a framework for generalizing \textbf{Conv}ersational dense retrieval via LLM-cognition data \textbf{Aug}mentation (\ours{}). 
We first generate multi-level augmented conversations to capture the diverse nature of conversational contexts.
Inspired by human cognition, we devise a cognition-aware prompting process to mitigate the generation of false positives, false negatives, and hallucinations.
Moreover, we develop a difficulty-adaptive sample filter that selects challenging samples for complex conversations, thereby giving the model a larger learning space.
A contrastive learning objective is then employed to train a better conversational context encoder.
Extensive experiments conducted on four public datasets, under both normal and zero-shot settings, demonstrate the effectiveness, generalizability, and applicability of \ours{}.
The code is released at \textcolor{magenta}{\url{https://github.com/haon-chen/ConvAug}}.
\end{abstract}

\section{Introduction}
Conversational search is anticipated to become the leading form of ad-hoc search engines in the future~\cite{cis_future}. 
This approach, utilizing multi-turn natural language interactions, offers a user-friendly experience, particularly for complex information-seeking tasks. 

There are two typical approaches for conversational search.
One way is conversational query rewriting (CQR)~\cite{acl23_ConvGQR, emnlp22_conqrr}. 
CQR models convert a conversational query into a de-contextualized search query suitable for ad-hoc retrieval.
However, CQR models either perform poorly because they cannot be optimized by downstream retrieval task~\cite{lecore}, or have unacceptable search latency when using large language models (LLMs) during inference~\cite{llmcs}.
Another approach is to perform conversational dense retrieval (CDR) in an end-to-end manner.
It typically uses the entire conversational context to train the context encoder within CDR models for passage retrieval.
This approach has been demonstrated to be more effective than CQR models on the downstream retrieval task of conversational search~\cite{emnlp23_instructor, lecore}.

Existing CDR approaches typically utilize conversations as fixed multi-turn natural language texts to train the context encoder.
However, in real-world scenarios, users can express conversations in various ways. 
The conversational search data often lack the diversity to support training for such variability due to the severe data sparsity issue. 
In other words, numerous alternative conversations with the same intent (or with similar expressions but different intents) as a specific data sample are unrecorded.
As a result, CDR models trained on these limited and fixed data often struggle to adapt to diverse real-world conversations.
Some works have tried to compensate for the deficiency of multi-turn texts.
However, these efforts often rely on basic rule-based strategies~\cite{coca} or human annotations to augment conversations~\cite{sigir22_COTED}.
Furthermore, comprehending turn dependencies in multi-turn conversations poses a significant challenge for simple language models.
 
To tackle these problems, we propose an LLM-based data augmentation framework to mimic how users perform diverse conversations.
Specifically, we design multi-level augmentation strategies to generate positive (similar intents but different expressions, denoted as $\pmb{+}$) and hard negative conversations (similar expressions but different intents, denoted as $\pmb{-}$):
(1) Token level. To mitigate the model's overreliance on specific tokens, we randomly mask some tokens of conversations ($\pmb{+}$).
Besides, we identify and replace the entities ($\pmb{-}$) to help the model focus on key information.
(2) Turn level. To prevent the model from depending on specific turns or the order of turns within conversations, we mask ($\pmb{+}$) and reorder ($\pmb{+}$) turns to generate diverse conversations.
We also generate a noisy turn ($\pmb{+}$) to enhance the model's denoising ability.
To avoid generating false positives, we identify the turn dependency structure to guide the turn-level augmentations.
(3) Conversation level. We paraphrase the conversation ($\pmb{+}$) to introduce linguistic variations.
We also shift the intent of conversations to help the model detect subtle intent changes ($\pmb{-}$).


However, LLMs may generate false positives or negatives and be prone to generate texts with hallucinations~\cite{HaluEval}.
To produce high-quality conversations, we propose a three-step prompting process inspired by human cognition.
Initially, we prompt an LLM to get a comprehensive understanding of the conversation (\eg, its intent and theme) in the first step~\cite{cognitive_stage1}.
Subsequently, the LLM associates existing elements, such as expressions, intents, and entities, with new yet related ones~\cite{cognitive_stage2}.
Finally, the LLM can conclude final outputs based on former outputs.
These outputs are less prone to be false positives, false negatives, or hallucinations, as the LLM has a deeper understanding of the original conversation (Step 1) and the generated elements are associated based on existing ones (Step 2).

Subsequently, we employ contrastive learning to bring together augmented positive conversations and push them away from negative ones.
Through this, we aim to train a more robust and generalized conversational context encoder, capable of accurately interpreting users' search intents of diverse conversations.
To enhance the contrastive learning process, we go beyond basic random sampling methods~\cite{coca}, and introduce a difficulty-adaptive sample filter to select more challenging augmented samples for more difficult conversations.
We believe that complex conversations offer a larger learning space for the model. 
More challenging data can thus provide the model with richer information, enabling it to understand these complex conversations better.

Extensive experiments on four public datasets demonstrate that \ours{} can consistently improve the performance of various conversational dense retrievers across various complexity levels of conversational turns.

The contributions of our work are as follows:

(1) We propose an LLM-based multi-level data augmentation framework \ours{} for conversational search.
It manages to comprehensively improve the effectiveness and generalizability of conversational retrievers.

(2) To obtain high-quality data, a cognition-aware prompting process is designed to prevent false positives/negatives and mitigate the hallucination problem of LLMs in conversation generation.

(3) We develop a difficulty-adaptive sample filter to select challenging samples for complex conversations to improve the model's understanding of those with large learning spaces.

\section{Related Work}

\noindent \textbf{Conversational search.}
CQR models usually utilize the context to rewrite the conversation into a standalone query~\cite{arxiv20_t5rewriter, emnlp22_CRDR, acl23_ConvGQR}.
Some researchers attempt to connect the downstream retrieval task to the rewriting task~\cite{emnlp22_conqrr, emnlp22_RLCQR, acl23_ediqr}.
On the other hand, CDR models try to utilize the whole conversation to train a conversational context encoder.
Some works use a few-shot manner to train the CDR model~\cite{sigir21_ConvDR, emnlp22_ConvTrans, arxiv24_haconvdr}.
Some design delicate denoising approaches for better CDR models~\cite{sigir22_COTED, mokdd23_preturns, lecore}.
However, none of these models focus on developing a context encoder that can comprehend diverse conversations smoothly.  

\noindent \textbf{Data augmentation for Information Retrieval.}
Because of the limited nature of relevance judgments, researchers of Information Retrieval (IR)~\cite{llmir_survey, sigir20_taggnn, add1, add2} have resorted to data augmentation.
Some use LLMs to generate queries from a
document~\cite{dair1}, or documents from a query~\cite{dair2, dair3, dair4} in ad-hoc retrieval.
For multi-turn ranking, some use basic rule-based approaches to generate variance of sequences for session search~\cite{coca}, personalized search~\cite{pssl}, and product search~\cite{cl_pps}.
COTED~\cite{sigir22_COTED} generates conversations based on human-annotated necessary historical turns. 

\noindent \textbf{LLM for Information Retrieval.}
LLMs have been widely used in various modules of the IR pipeline~\cite{LLM4IRSurvey}, such as retriever~\cite{llmir_tart}, reranker~\cite{llmir_rankllama}, and reader~\cite{llmir_self-RAG}.
In conversational search, some employ LLMs to aid the training~\cite{emnlp23_llmaided, interpretingCDR} and the inference~\cite{llmcs} stage of CQR.
Instructor~\cite{emnlp23_instructor} uses LLMs to generate pseudo passage labels to facilitate unsupervised CDR models.
However, these models fail to utilize LLMs to alternate the contexts for a generalized context encoder.
\section{Methodology: \ours{}}

\begin{figure}[!t]
	\centering
	\includegraphics[width=1.0\linewidth]{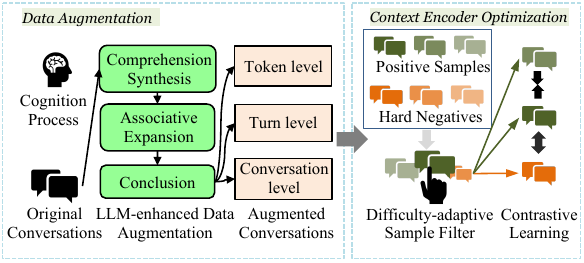}
	\caption{The training workflow of our framework. }
	\vspace{-2ex}
	\label{fig:workflow}
\end{figure}

In this section, we present our two-stage framework \ours{}, 
as illustrated in Figure~\ref{fig:workflow}.
In the first stage, we leverage an LLM to perform our data augmentation strategies tailored for conversational search. 
A three-step cognition-aware prompting process is developed to guide the LLM to generate multi-level augmented conversations.
The second stage is to utilize the augmented data to optimize the conversational context encoder.
We propose to select more challenging samples for more complex conversations to facilitate model learning.

\subsection{Problem Formulation}
In this work, we focus on the conversational passage retrieval task.
The context of a conversation is denoted as $ C_n = \{q_1, r_1, ..., q_{n-1}, r_{n-1}, q_n\}$, where $q_i$ and $r_i$ are the query and response of the $i$-th turn ($t_i$) in $C_n$, and $q_n$ is the current query.
Given $C_n$, our goal is to retrieve the relevant passage $d^+$ from the passage collection $\mathcal{D}$.
For convenience, we will omit the subscript $n$ in the rest of this paper.

\subsection{LLM-enhanced Data Augmentation} \label{subsec:llm_dataaug}

\begin{figure*}[!t]
	\centering
	\includegraphics[width=0.95\textwidth]{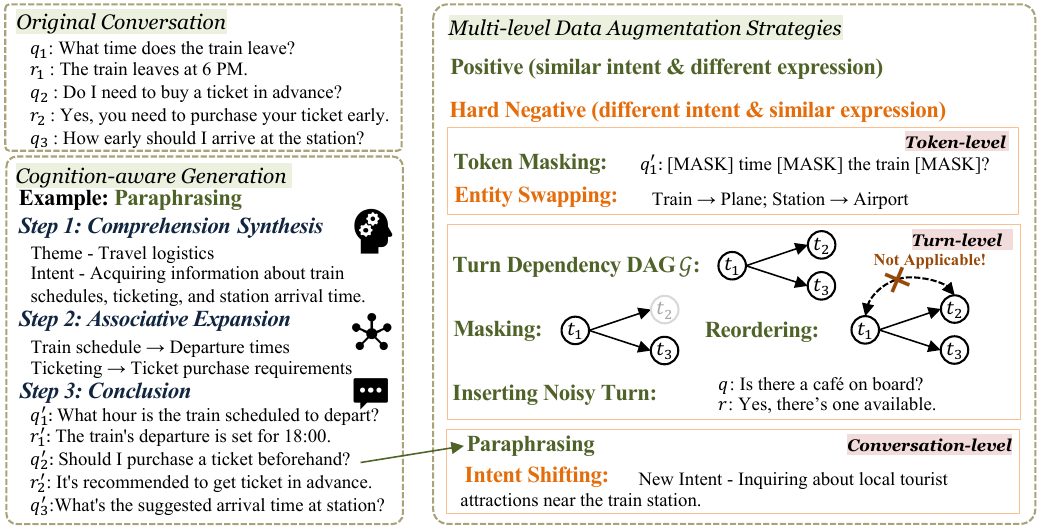}
	\caption{An example to illustrate our cognition-aware prompting process and multi-level augmented data. }
	\vspace{-2ex}
	\label{fig:data_aug}
\end{figure*}

Conversational search suffers from a severe data sparsity issue, \ie, varying expressions of recorded conversations are inadequate, leading to insufficient training of context encoders.
As shown in Figure~\ref{fig:data_aug}, we propose to mimic the diverse ways users might express conversations by developing data augmentation strategies.
We propose both positive ($\pmb{+}$) and hard negative ($\pmb{-}$) generation strategies to produce conversations with similar ($C^+$) and different intents ($C^-$), respectively.
Furthermore, the LLM-based generation is prompted by a three-step cognition-aware process to mitigate hallucinations and enhance the data quality.

\subsubsection{Multi-level Conversation Alteration} \label{subsubsec:multilevel_aug_convs}

 \noindent $\bullet$ \textbf{Token-level alteration} 
 
Firstly, we propose to perform fine-grained token-level alterations on $C$ to help the model learn nuanced information.

\emph{Token Masking} ($\pmb{+}$). To prevent the model from relying too much on specific tokens, we employ a rule-based strategy.
A context is treated as a sequence of tokens: $C= \{w_1, w_2, \ldots, w_{M}\}$, where $M$ is the total number of tokens.
We randomly mask a proportion $r_\text{w}$ of the tokens in $C$ with a special token ``[token\_mask]''.
By this, we aim to produce a similar context $C^+_{\text{tom}}$ as it only has little differences from $C$ in some tokens.

\emph{Entity Replacing} ($\pmb{-}$). In real-world scenarios, the same conversational flow can occur with different entities.
We use the LLM to identify and replace entities in $C$ to generate $C^-_{\text{ent}}$, which is contextually similar to $C$ but differs in critical details.
By contrasting it to other $C^+$, the model can pay closer attention to the key information in the context rather than the superficial aspects. 

\noindent $\bullet$ \textbf{ Dependecy-aware turn-level alteration}

Secondly, we propose more coarse-grained alterations at the turn level.
As shown in Figure~\ref{fig:data_aug}, the understanding of $t_2 = (q_2, r_2)$ and $t_3 = (q_3)$ both depend on $t_1$ since they all need the information ``train''.
Therefore, the dependencies within conversations are important if we want to alternate them without changing their search intents, \ie, avoiding producing false positives.
Utilizing the ability of LLMs, we can identify the necessary historical turns of $t_i$ automatically. 
After performing this sequentially on all turns of $C$, we can construct a {turn dependency Directed Acyclic Graph (DAG)} $\mathcal{G}$, as shown in the right part of Figure~\ref{fig:data_aug}.

\emph{Turn Masking} ($\pmb{+}$). For all historical turns $T_\text{h} = \{t_1, t_2, \ldots, t_{n-1}\}$ of $C$, we mask a proportion $r_\text{t}$ of the turns with a special token ``[turn\_mask]'' to generate $C^+_{\text{tum}}$.
With this, \ours{} is forced to not rely on specific turns and get a more robust understanding of the whole conversation.
To maintain the dependency structure of $C$, we can only mask the turns that are {not the ancestors} of $t$.

\emph{Turn Reordering} ($\pmb{+}$). We select a pair of historical turns $(t_i, t_j)$ in $T_\text{h}$ and swap their positions to produce $C^+_{\text{reo}}$.
We can only choose turns that the {topological ordering of $\mathcal{G}$ remains the same} after the swapping.
Through this restriction, $C^+_{\text{reo}}$ will have a different order of expression while maintaining the logical chain as $C$.
This process challenges the model to focus more on the content of each turn rather than just the order.

\emph{Inserting Noisy Turn} ($\pmb{+}$). Conversations are often interrupted by unrelated interjections.
Corrupting the current context can help the model handle conversational dynamics.
We extend the existing context for one additional noisy turn $t_{\text{noi}}$ and randomly insert it into $T_\text{h}$.
Since we prompt the LLM to generate a turn that is relevant to the main background of $C$ but introduces a slightly divergent element, the generated turn can be inserted into any position in $T_\text{h}$ to produce $C^+_{\text{noi}}$ without disrupting the dependency structure. 

\noindent $\bullet$ \textbf{Conversation-level alteration}

At last, we apply more high-level changes to the whole conversation.

\emph{Paraphrasing} ($\pmb{+}$). To mimic users' various expressions of similar intents, we aim to use the LLM to expand the linguistic diversity by paraphrasing the whole $C$ to produce $C^+_{\text{para}}$.
This can help reduce the model's tendency to overfit specific phrasings or patterns of $C$, which enhances the model's ability to generalize to unseen conversations.

\emph{Intent Shifting} ($\pmb{-}$). The intent behind a dialogue can shift subtly without significant changes in the expression of the conversation.
Therefore, we utilize the LLM to produce the intent-shifted conversations $C^-_{\text{int}}$.
By contrasting them to $C^+$, our model is trained to detect and adapt to subtle intent shifts in real conversations.

\subsubsection{Cognition-aware Prompting Process} \label{subsubsec:cog_gen}


To enhance the data quality, we propose a three-step prompting method inspired by human cognition theory, including \textit{Comprehension Synthesis (Step 1)}, \textit{Associative Expansion (Step 2)}, and \textit{Conclusion (Step 3)}.
As shown in Figure~\ref{fig:data_aug}, we take the paraphrasing strategy as an example for illustration:

\noindent \textbf{Step 1: Comprehension Synthesis.} 
When we have a conversation, our brains initially construct a comprehensive representation of the text~\cite{cognitive_stage1}. 
This step allows the LLM to have a comprehensive understanding of the whole conversation.
Specifically, we prompt the LLM using "\textit{Step 1: Comprehension Synthesis: [Identify key themes and
intents of the conversation]}".
The understanding of these core aspects will prevent the LLM from generating $C^+_{\text{para}}$ that deviates from the theme and search intents (false positive).

\noindent \textbf{Step 2: Associative Expansion.}
The human mind often uses spreading activation in semantic networks, where one concept triggers related concepts~\cite{cognitive_stage2}.
Inspired by this theory, the prompt we give the LLM is "\textit{Step 2: Associative Expansion: [Generate alternative expressions based on existing ones]}".
This step serves as an intermediate process that leverages LLM's creativity to think of novel elements while preventing it from hallucinating unrelated elements.

\noindent \textbf{Step 3: Conclusion.}
In the final step, we prompt the LLM as: "\textit{Step 3: Conclusion: [Paraphrase the conversation based on outputs of last two steps]}".
In our example, the output is a paraphrased conversation that maintains $C$'s search intent (Step 1) while introducing new but related (Step 2) expressions, avoiding false positives and hallucinations.

We manually write several demonstrations for each step to prompt an LLM to do in-context generation.
The complete prompts are in Appendix~\ref{appendix: prompt}.

\subsection{Training Conversational Context Encoder} \label{subsec:CDR_train}

Through our proposed data augmentation strategies, we can generate a set of positive samples $\mathcal{C}^+ = \{C^+_{\text{tom}}, C^+_{\text{tum}}, C^+_{\text{reo}}, C^+_{\text{noi}}, C^+_{\text{para}}\}$ and hard negative samples $\mathcal{C}^- = \{C^-_{\text{ent}}, C^-_{\text{int}} \}$ for an original conversation $C$ in the dataset.
Then, to enhance model learning, we develop a difficulty-adaptive sample filter to keep samples of matching difficulty for original conversations.
Finally, we train the conversational context encoder on these augmented samples with multi-task contrastive learning.

\subsubsection{Difficulty-adaptive Sample Filter} \label{subsubsec:DSA}

Considering that simple augmentations for complex $C$ may result in underfitting, and complex augmentations for simple $C$ can cause overfitting, we develop a difficulty-adaptive sample filter. 
It selects difficult samples for difficult conversations to enhance the training process.

Specifically, the difficulty of the original conversations is defined as:
$\text{Diff}(C) = |T_\text{h}|+\left(|\text{Topic}(C)|*\overline{\text{PPL}(C)}\right) $,
where $|T_\text{h}|$ denotes the number of the historical turns, $|\text{Topic}(C)|$ is the number of topics , and $\overline{\text{PPL}(C)}$ denotes the average perplexity of $C$.
The detailed calculation of these components can be found in Appendix~\ref{sec:topic_count}.
To give the diversity of topics and the linguistic challenges more emphasis, we compute $|\text{Topic}(C)|*\overline{\text{PPL}(C)} $ and use $|T_\text{h}|$ as an indicator of rich information within long conversations.

%
\begin{figure}[!t]
	\centering
	\includegraphics[width=0.95\linewidth]{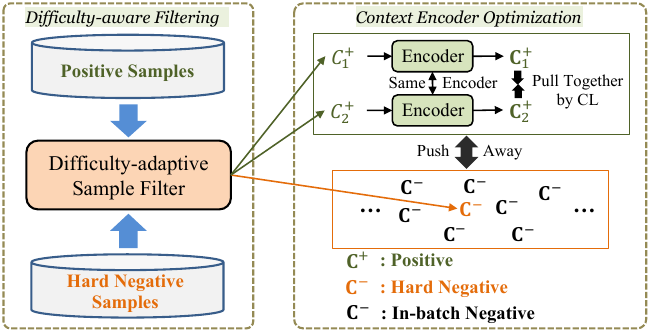}
	\caption{The optimization of context encoder. }
	\vspace{-2ex}
	\label{fig:optimize}
\end{figure}
%

For the difficulty of the augmented conversations,
we first obtain paired positive samples: $\mathcal{P}_{\mathcal{C}^+} = \{ (C_i^+, C_j^+) \mid C_i^+, C_j^+ \in \mathcal{C}^+, i \neq j \}$.
We then use a sentence-transformers model to compute the similarity of each pair, the difficulty is denoted as $\text{Diff}^+(C_i^+, C_j^+) = 1-\text{BERTSim}(C_i^+, C_j^+)$, where BERTSim($\cdot$) is the cosine similarity of encoded conversations.

For the diversity of used augmented samples, we divide all training conversations into $|\mathcal{P}_{\mathcal{C}^+}|$ buckets based on $\text{Diff}(C)$.
We then filter and select one positive pair with matching $\text{Diff}^+(C_i^+, C_j^+)$ for each conversation.
As for hard negatives, we pair each negative with selected positive samples: $\text{Diff}^-(C_h^-) = (\text{BERTSim}(C_i^+, C_h^-)+\text{BERTSim}(C_j^+, C_h^-))/2$. We select $k$ hard negatives with higher $\text{Diff}^-(C_h^-)$ for difficult $C$.

\subsubsection{Multi-task Contrastive Learning} \label{subsubsec:CL_train}

For the ranking task, we apply a standard ranking loss based on contrastive learning of passages: 
\begin{equation}
    \mathcal{L}_{\text{rank}} = -\log\frac{e^{(\mathbf{C} \cdot \mathbf{d}^+)}}{e^{(\mathbf{C} \cdot \mathbf{d}^+)}+\sum_{{d}^-\in\mathcal{D}}{e^{(\mathbf{C} \cdot \mathbf{d}^-)}}},
\end{equation}
where $\mathbf{C}=\text{CCE}(s)$ denotes $C$ encoded by the conversational context encoder and $s = \text{[CLS]}\circ q_n \circ r_{n-1} \circ \ldots \circ r_1 \circ q_1 \circ \text{[SEP]}$ is the concatenated sequence of $C$.
$\mathbf{d}^+$ and $\mathbf{d}^-$ are encoded by the frozen passage encoder $\mathbf{d} = \text{PE}({d})$.

Suppose a minibatch contains $N$ conversations, we use our difficulty-adaptive sample filter to select two positive samples for each $C$ to form $\{\mathcal{X}\}$ comprising $2N$ sequences.
The two sequences derived from the same $C$ are considered a similar pair, whereas the remaining $2(N-1)$ serve as in-batch negative samples $\{\mathcal{X}\}^-$.
Besides, we select $k$ hard negative samples for each $C$ to form $\{\mathcal{H}\}$ comprising $kN$ sequences.
The contrastive learning loss for a positive pair $({C}_i^+,{C}_j^+)$ and negatives ${C}^- \in \{\{\mathcal{X}\}^-\cup \mathcal{H}\}$ of $C$ is formulated as follows:
\begin{eqnarray}
\mathcal{L}_{\text{CL}}(i,j) = -\log\frac{\phi(\mathbf{C}_i^+,\mathbf{C}_j^+)}{\phi(\mathbf{C}_i^+,\mathbf{C}_j^+) + \sum\limits_{}^{}\phi(\mathbf{C}_i^+,\mathbf{C}^-)}
\end{eqnarray}
where $\phi(\cdot) = \exp (\text{cos}(\cdot)/\tau)$, ${\rm cos}(\cdot)$ is cosine similarity and $\tau$ is a hyperparameter temperature.

We optimize these two tasks together as: $\mathcal{L} = \mathcal{L}_{\text{rank}} + \alpha \mathcal{L}_{\text{CL}}$, where $\alpha$ is used to balance losses.
\section{Experiments}

\begin{table*}[!t]
\centering
\scalebox{0.94}{\begin{tabular}{c|c|ccc|ccc}
\toprule
\multirow{2}{*}{Category} & \multirow{2}{*}{Model}   & \multicolumn{3}{c|}{QReCC}                    & \multicolumn{3}{c}{TopiOCQA}         \\ \cline{3-8} 
                          &                          & MRR           & NDCG@3        & Recall@10     & MRR        & NDCG@3     & Recall@10  \\ \midrule
\multirow{4}{*}{CQR Models}      & T5QR                     & 34.5          & 31.8          & 53.1          & 23.0       & 22.2       & 37.6       \\
                          & ConQRR                   & 41.8          & -             & 65.1          & -          & -          & -          \\
                          & ConvGQR                  & 42.0          & 41.0          & 64.4          & 25.6       & 24.3       & 41.8       \\
                          & ED                       & 49.4          & -             & 67.0          & -          & -          & -          \\ \midrule
\multirow{6}{*}{CDR Models}      & ConvDR                   & 38.5          & 35.7          & 58.2          & 27.2       & 26.4       & 43.5       \\
                          & InstructoR-ANCE               & 43.5          & 40.5          & 66.7          & 25.3       & 23.7       & 45.1       \\
                          & Conv-ANCE                & 49.0          & 46.6          & 71.4          & 30.4       & 28.5       & 52.6       \\
                          & Conv-SPLADE              & 50.0          & 46.6          & 69.9          & 30.7       & 29.5       & 52.1       \\
                          & LeCoRE                   & {\ul 51.1}    & {\ul 48.5}    & {\ul 73.9}    & {\ul 32.0} & {\ul 31.4} & {\ul 54.3} \\ 
                          & \ours{} (Ours) & \textbf{52.7$^\dag$} & \textbf{50.4$^\dag$} & \textbf{75.6$^\dag$} & \textbf{35.0$^\dag$}  & \textbf{33.3$^\dag$}  & \textbf{57.9$^\dag$}  \\ \bottomrule
\end{tabular}}
\caption{The results of the normal evaluation.
``$\dag$'' denotes our model outperforms all baselines significantly except CONQRR and ED.
The best performance is in bold and the second-best performance is underlined.}
\label{tab:result}
\end{table*}

\subsection{Datasets and Metrics}
We evaluate our model with both normal and zero-shot evaluation.
Following previous CDR works~\cite{lecore, emnlp23_instructor}, we train \ours{} on \textbf{QReCC}~\cite{naacl21_qrecc} and \textbf{TopiOCQA}~\cite{tacl22_topiocqa2}.
Additionally, we test \ours{} that has been trained on QReCC in a zero-shot setting on \textbf{CAsT-20}~\cite{trec_cast20} and \textbf{CAsT-21}~\cite{trec_cast21}.
We omit the CAsT-19 dataset since it is less challenging and realistic compared to CAsT-20 and CAsT-21~\cite{llmcs}.
More details are in Appendix~\ref{appendix: dataset}.

Following previous works~\cite{emnlp23_llmaided}, we use some popular metrics for normal evaluation: MRR, NDCG@3, Recall@10.
For zero-shot setting, we use metrics suggested by CAsT~\cite{trec_cast21}: MRR, NDCG@3.
All significant tests are done using paired t-tests at $p<0.05$ level with Bonferroni correction.

\subsection{Implementation Details}
We adopt ANCE~\cite{iclr21_ance} as the base model of \ours{}.
For the large language model, we use Llama 2-Chat (7B)~\cite{llama2} to perform our data augmentation tasks.
We use $k=1$ augmented negative conversations as hard negatives.
More details about training and hyperparameters are in our code and Appendix~\ref{appendix: implementation}.

\subsection{Baselines}
We compare \ours{} with two kinds of models:

\noindent \textbf{Conversational query rewriter.}
$\bullet$~T5QR~\cite{arxiv20_t5rewriter}  trains the rewriter with the human rewrites.
$\bullet$~ConQRR~\cite{emnlp22_conqrr} employs reinforcement learning to train CQR models.
$\bullet$~ConvGQR~\cite{acl23_ConvGQR} reformulates better conversational queries by relating to the retrieval task. 
$\bullet$~ED~\cite{emnlp23_llmaided} distills the rewriting capabilities of LLMs into smaller models.
Note we do not compare those using black-boxed LLMs (\eg, ChatGPT) during inference ~\cite{llmcs} since these models require significant resources and time to invoke API numerous times during inference.

\noindent \textbf{Conversational dense retriever.}
$\bullet$~ConvDR~\cite{sigir21_ConvDR} distills knowledge for few-shot learning.
$\bullet$~Conv-ANCE~\cite{arxiv20_t5rewriter} \& Conv-SPLADE~\cite{sigir21_SPLADE} are ANCE and SPLADE fine-tuned on the training conversations with only the training loss.
$\bullet$~ConvDR~\cite{sigir21_ConvDR} distills knowledge for few-shot learning.
$\bullet$~LeCoRE~\cite{lecore} extends SPLADE by generating denoised and interpretable lexical session representation.
$\bullet$~InstructoR~\cite{emnlp23_instructor} employs LLMs to estimate the session-passage relevance score to guide the retriever’s training.
We use the ``ANCE+$\text{InstructoR}_{\text{QRPG}}$'' version for fair comparisons with \ours{}.

\subsection{Overall Results}

\subsubsection{Normal Evaluation}

We conduct the normal evaluation on QReCC and TopiOCQA, and the results are presented in Table~\ref{tab:result}. We can make these observations:
(1) \ours{} outperforms all baseline models significantly on both datasets.
This demonstrates the effectiveness of our LLM-enhanced data augmentation and context encoder optimization.
Furthermore, based on the model ANCE, whose performance is comparable to SPLADE, \ours{} still manages to gain superior performance than the SPLADE-based model LeCoRE.
This further indicates that our approach can train a more robust and generalized context encoder.
(2) CDR models generally outperform CQR models.
We can observe that even the simply fine-tuned model Conv-ANCE still outperforms the LLM-aided CQR model ED.
This indicates the importance of the ranking signal and the effectiveness of our multi-task learning approach.

\subsubsection{Zero-shot Evaluation}

\begin{table}[t!]
\centering
\small
\setlength{\tabcolsep}{2pt}
\begin{tabular}{c|cc|cc}
\toprule
\multirow{2}{*}{Model}   & \multicolumn{2}{c|}{CAsT-20}  & \multicolumn{2}{c}{CAsT-21}   \\ \cline{2-5} 
                         & MRR           & NDCG@3        & MRR           & NDCG@3        \\ \hline
InstructoR-ANCE               & {\ul 43.7}    & {\ul 29.6}    &  {\ul 53.0}    & {\ul 34.9}             \\
Conv-ANCE                & 42.2          & 27.7          & 52.3          & 34.2          \\
Conv-SPLADE              & 36.9          & 28.1          & 47.9          & 29.9          \\
LeCoRE                   & 37.7          & 29.0          & 50.8          & 32.3          \\
\ours{} (Ours) & \textbf{45.0}$^\dag$ & \textbf{30.7}$^\dag$ & \textbf{54.8}$^\dag$ & \textbf{36.8}$^\dag$ \\
\bottomrule
\end{tabular}
\caption{The performances of CDR models at zero-shot setting.
``$\dag$'' denotes our model outperforms all baselines significantly.
The best performance is in bold and the second-best performance is underlined.}
\label{tab:zero-shot}
\end{table}

We also evaluate our model's generalization ability by conducting a zero-shot test of CDR models trained on QReCC on two challenging datasets CAsT-20 and CAsT-21. 
From the results in Table~\ref{tab:zero-shot}, we can make the following observations:
(1) \ours{} consistently outperforms all CDR baseline models in terms of both metrics on all datasets.
Specifically, \ours{} maintains its superiority over ANCE-based CDR models (Conv-ANCE and InstructoR-ANCE), which further demonstrates the generalization ability of \ours{}.
(2) The unsupervised model InstructoR-ANCE gains the second-best performance in the zero-shot setting. 
For example, it gains a performance of 43.7 in terms of MRR on CAsT-20.
However, its performance is poor in the normal setting.
This indicates that this unsupervised approach might not align well with labeled tasks but it can be effectively applied to unseen datasets.

\subsection{Ablation Study}

\begin{table}[t!]
    \centering
    \small
    \begin{tabular}{p{0.25\textwidth}cc}
    \toprule
         Model & {MRR} & {NDCG@3}  \\
        \midrule
        \ours{} (Full) & \textbf{52.7}$^\dag$ & \textbf{50.4}$^\dag$\\
        \quad w/o. Token Masking ($C^+_{\text{tom}}$) & 51.2  & 48.9   \\
        \quad w/o. Turn Masking ($C^+_{\text{tum}}$)& 51.9  & 49.6   \\
        \quad w/o. Turn Reordering ($C^+_{\text{reo}}$)& 52.0  & 49.5   \\
        \quad w/o. Noisy Turn ($C^+_{\text{noi}}$)& 52.3  & 49.9   \\
        \quad w/o. Dependency-aware & 52.0  & 49.6  \\
        \quad w/o. Paraphrasing ($C^+_{\text{para}}$)& 52.1 & 49.8   \\
        \quad w/o. Entity Replacing ($C^-_{\text{ent}}$)& 50.8  & 48.5   \\
        \quad w/o. Intent Shifting ($C^-_{\text{int}}$)& 52.4  & 50.0   \\
        \midrule
        \quad w/o. Cognition-aware & 51.1  & 49.0    \\
        \midrule
        \quad w/o. Filter (rand) & 51.7  & 49.5   \\
        \quad w/o. Filter (easy) & 51.6  & 49.3  \\
    \bottomrule
    \end{tabular}
    \caption{Performances of ablated models on QReCC.
    ``$\dag$'' denotes \ours{} outperforms ablated models significantly.}
    \label{tab:ablation}
\end{table} 

To evaluate the effectiveness of each component, we conduct ablation studies on \ours{}:

\noindent \textbf{Data augmentation strategies.}
We first conduct ablation experiments on our multi-level data augmentation strategies.
As shown in Table~\ref{tab:ablation}, the performance of \ours{} drops significantly after discarding each kind of alteration.
Specifically, the performance of \ours{} drops most when we discard the strategy Entity Replacing ($C^-_{\text{ent}}$).
This demonstrates that teaching our model to pay more attention to key information in conversations is effective for understanding search intents.
Additionally, we find that \ours{}'s performance decreases if we do not mask or reorder turns based on the dependency graph constructed by the LLM.
All these results demonstrate the effectiveness of our designed data augmentation strategies.

\noindent \textbf{Cognition-aware prompting process.}
We also replace the three-step prompting process with a naive prompt template (Appendix~\ref{subsec:other_prompt}) and train ``\ours{} w/o. Cognition-aware'' on data generated by this prompt.
The performance of \ours{} decreases by about 3\% in terms of MRR when we replace the cognition prompting process.
This demonstrates that our cognition-aware prompting process can produce data with higher quality.

\noindent \textbf{Difficulty-adaptive sample filter.}
We replace our filter with a random selector (\ours{} w/o. Filter (rand)) and one that selects easy samples for difficult conversations (\ours{} w/o. Filter (easy)).
The decrease in \ours{}'s performance demonstrates that selecting challenging augmented samples for difficult conversations can help the model understand them better.
Specifically, the performance of \ours{} decreases if we assign easy samples to difficult conversations (even worse than randomly selecting).
This further demonstrates that we will underfit \ours{} if we do not give harder conversations enough learning space.

\subsection{Performance on Different Turns}

\begin{figure*}[!t]
	\centering
	\includegraphics[width=0.9\textwidth]{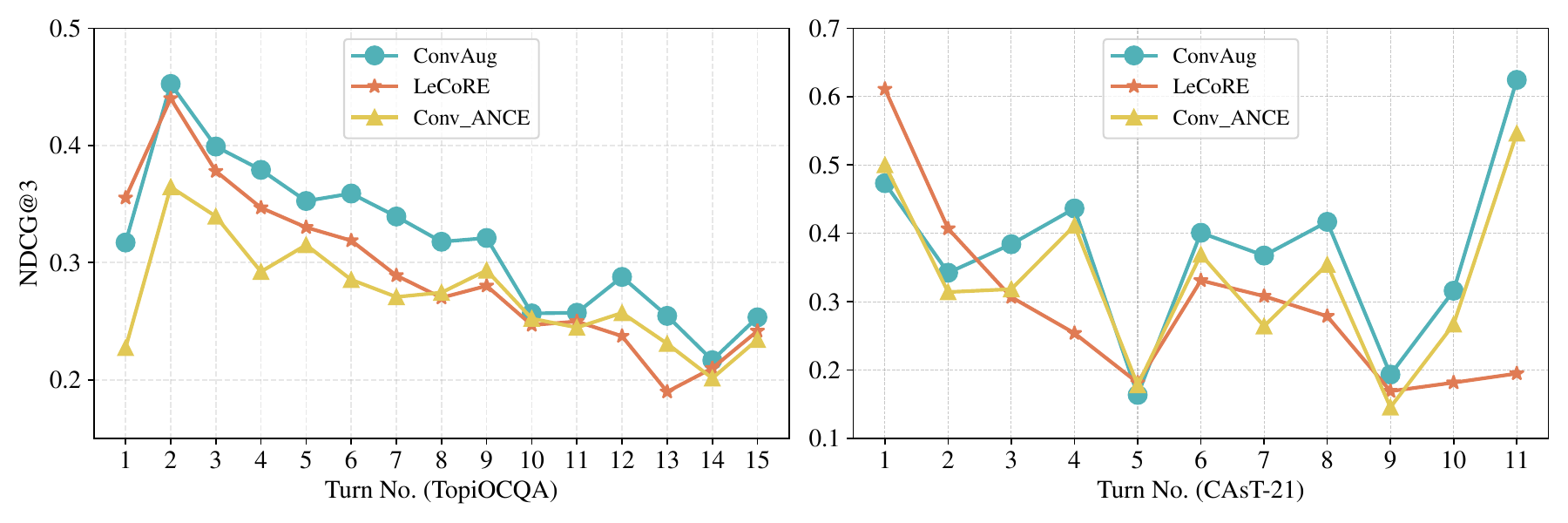}
	\caption{Turn-level performance comparisons on TopiOCQA (normal) and CAsT-21 (zero-shot). }
	\vspace{-2ex}
	\label{fig:turn}
\end{figure*}

To investigate the performance of \ours{} at a more fine-grained level, we compare it with LeCoRE and Conv-ANCE at the turn level using TopiOCQA (normal) and CAsT-21 (zero-shot) datasets.
The results, as shown in Figure~\ref{fig:turn}, indicate that \ours{} surpasses both baselines in the majority of turns, underscoring its effectiveness and generalizability again. 
Specifically, \ours{} shows more significant improvements in later conversation turns (\eg, from the 2nd to the 15th turns on TopiOCQA and the 3rd to the 11th turns on CAsT-21). 
This is because longer conversations often contain more diverse information and our augmented data can help \ours{} to generalize to these complex conversations.
Besides, our difficulty-adaptive sample filter can challenge \ours{} to learn more about complex conversations.

\subsection{Influence of Augmented Hard Negatives}

\begin{table}[t!]
    \centering
    \small
    \begin{tabular}{l|cc|cc}
    \toprule
\multirow{2}{*}{Ratio}   & \multicolumn{2}{c|}{QReCC}  & \multicolumn{2}{c}{CAsT-21}   \\ \cline{2-5} 
                         & MRR           & NDCG@3        & MRR           & NDCG@3        \\ \hline
$k=0$              & {50.8}    & {48.4}    & 53.3        &  35.3             \\
$k=1$              & \textbf{52.7}$^\dag$         & \textbf{50.4}$^\dag$          & \textbf{54.8}$^\dag$      & \textbf{36.8}$^\dag$          \\
$k=2$              & 51.5          & 49.0          & 50.8         & 34.3       \\
\bottomrule
    \end{tabular}
    \caption{ Performances of \ours{} with different ratios $k$ of hard negative samples.
    ``$\dag$'' indicates the result is significantly better than others. } 
    \label{tab:hard_neg}
\end{table}

We use $k$ generated hard negative contexts to facilitate the training of \ours{}'s context encoder.
The performances of \ours{} with different $k$s are in Table~\ref{tab:hard_neg}.
We can observe that \ours{} performs best on QReCC with $k=1$ hard negative.
We believe there is a trade-off.
The lack of hard negatives limits the model's ability to benefit from challenging comparisons, leading to a less robust feature representation. 
On the other hand, incorporating multiple hard negatives may introduce noise or ambiguity, potentially corrupting the learning process.
Besides, we can observe that \ours{} ($k=0$) performs better on zero-shot than on normal evaluation.
This further demonstrates that too many hard negative samples will introduce noise and harm the model's generalizability.

\subsection{Application to Other Base Retrivers}

\begin{table}[t!]
    \centering
    \small
    \begin{tabular}{lccc}
    \toprule

        \textbf{Model} & MRR & NDCG@3 \\ 
        
        \midrule
        
        Conv-SPLADE & 50.0          & 46.6  \\
        
        Conv-SPLADE + \ours{} & \textbf{52.4}$^\dag$ & \textbf{49.8}$^\dag$  \\
        
        \midrule
        
        LeCoRE & {51.1}    & {48.5} \\
        
        LeCoRE + \ours & \textbf{53.1}$^\dag$ & \textbf{50.7}$^\dag$ \\
 
    \bottomrule
    \end{tabular}
    \caption{ The performances of the base models and the models with our training framework (\ours{}) on the QReCC dataset.
    ``$\dag$'' indicates the result in bold is significantly better than the base model. } 
    \label{tab:other_bases}
\end{table}

We use ANCE as the base model of \ours{} since it is a popular dense retriever that has been the base model of many CDR models.
However, our training framework can be easily applied to other CDR models.
We choose Conv-SPLADE and LeCoRE as the base models and apply our approach to them.
From the results shown in Table~\ref{tab:other_bases}, we can observe that our method can bring significant improvements across different base CDR models (even sparse retrievers).
This demonstrates the broad applicability of our approach.
\section{Conclusion}

In this work, we present \ours{} to augment conversational search data with LLMs.
We design a three-step cognition-aware prompting process to generate multi-level augmented conversations.
We also develop a difficulty-adaptive sample filter to assign challenging samples to difficult conversations for larger learning space.
A contrastive learning objective is employed to train a generalized conversational context encoder.
Extensive experiments on four public datasets at both normal and zero-shot settings validate the effectiveness, generalization ability, and applicability of \ours{}.

\section*{Limitations}
For future studies, our work has the following limitations that we plan to address:
\begin{enumerate}
    \item The equation we developed to assess the complexity of conversations is relatively basic. We plan to design a more sophisticated equation of our three components in the future.
    \item We use an LLM to augment the training conversations in the pre-processing stage. Although the inference time remains the same as base retrievers, the augmentation process takes quite a long time because of the data amount we need to generate (millions of conversations) and the limited computational resources (4 NVIDIA A100 GPUs).
    \item We only conduct experiments using one LLM Llama 2 (7B) due to the cost of augmenting such a large number of data. Performances of other LLMs will be experimented with in the future.
    \item There is also a potential risk involved.
    Since we are using LLMs to generate conversations, the original data should not contain sensitive or private information that may cause LLMs to produce risky texts.
\end{enumerate}

\section*{Acknowledgement}
This work was supported by the National Natural Science Foundation of China No. 62272467, the fund for building world-class universities (disciplines) of Renmin University of China, and Public Computing Cloud, Renmin University of China. The work was partially done at the Engineering Research Center of Next-Generation Intelligent Search and Recommendation, MOE.

\appendix

\clearpage

\section*{Appendix}

\section{Dataset Details}
\label{appendix: dataset}

In this part, we will introduce more details of the four datasets we use.

\textbf{QReCC} represents the large-scale, open-domain conversational question-answering (QA) dataset featuring human-annotated question rewrites. It integrates conversations from QuAC~\cite{emnlp18_QuAC}, TREC CAsT, and Natural Questions~\cite{tacl19_NQ}. The text corpus used for retrieval contains 54 million passages.

\textbf{TopiOCQA} comprises conversations coming from a real search query found in Natural Questions, with subsequent interactions simulated using a wizard-of-oz approach.

\begin{table}[h]
    \centering
    \begin{tabular}{lcc}
    \toprule
        \textbf{QReCC} & \textbf{Training} & \textbf{Testing} \\
    \midrule
        \# Conversations    & 10,823         & 2,775      \\
\# Turns            & 63,501        & 16,451     \\
\# Passages         & \multicolumn{2}{c}{54M}             \\
    \midrule
        \textbf{TopiOCQA} & \textbf{Training}  & {\textbf{Testing}} \\
    \midrule
        \# Conversations    & 3,509         & 205      \\
\# Turns            & 45,450        & 2,514     \\
\# Passages         & \multicolumn{2}{c}{25M}      \\
    \bottomrule
    \end{tabular}
    \caption{Statistics of QReCC and TopiOCQA.}
    \label{tab:qrecc_statistics}
\end{table}

\textbf{CAsT-20} and \textbf{CAsT-21} were released by the TREC Conversational Assistance Track (CAsT). 
Their limited number of conversations often makes them evaluation datasets.
Each query turn in both CAsT-20 and CAsT-21 has a corresponding human rewrite a canonical response passage.

\begin{table}[h]
\centering
\begin{tabular}{lcc}
\toprule
Dataset             & \textbf{CAsT-20}    & \textbf{CAsT-21} \\ \midrule
\# Conversations    & 25         & 18      \\
\# Turns            & 208        & 157     \\
\# Passages         & 38M        &  40M       \\ \bottomrule
\end{tabular}
\caption{Statistics of the {CAsT} datasets.}
\label{tab:cast_statistics}
\end{table}

\section{Implementation Details}
\label{appendix: implementation}

We use ANCE provided by Huggingface as the base model\footnote{\url{https://huggingface.co/castorini/ance-msmarco-passage}}.
We use $k=1$ augmented negative conversations as hard negative.
We set the temperatures as 0.0012 and 0.001 for training conversational context encoders on QReCC and TopiOCQA, respectively.
The token mask ratio $r_\text{w}$ and turn mask ratio $r_\text{t}$ are tuned and established as 0.5 and 0.5, respectively for the QReCC dataset and 0.9 and 0.5, respectively for the TopiOCQA dataset.
The learning rates are set as 1e-5 and 1.5e-5 for training on QReCC and TopiOCQA, respectively.
The weight $\alpha$ is set as 1.0 and 0.1 for QReCC and TopiOCQA, respectively.
The model is trained with a batch size of 12.
More details can be found in our code.

\section{Prompt Templates}
\label{appendix: prompt}

\subsection{Multi-level Data Augmentaion}
\label{subsec:data_aug_prompt}

\begin{prompt}[title={Prompt: Entity Replacing}, label=prompt:ent]
Task Overview:

Your task is to replace entities in the current conversation context while keeping the expressions as similar as possible to the original. This involves identifying key entities, replacing them with suitable alternatives, and ensuring the conversation remains coherent. Use the following structured approach:

Example to Illustrate the Process (\textcolor{blue}{Demonstration}):

Original Conversation:

Query1: "How long is the Golden Gate Bridge?"

Response1: "The Golden Gate Bridge is about 1.7 miles long."

Query2: "When was it opened to the public?"

Response2: "It was opened in May 1937."

\textcolor{blue}{Step 1}: Comprehension Synthesis (Identify key entities in the conversation)

Output: Key Entities - Golden Gate Bridge, 1.7 miles, May 1937.

\textcolor{blue}{Step 2}: Associative Expansion (Find suitable replacements for the identified entities)

Output: Brooklyn Bridge, 1.1 miles, December 1883.

\textcolor{blue}{Step 3}: Conclusion (Reconstruct the conversation with new entities)

Entity-replaced Conversation:

Query1: "How long is the Brooklyn Bridge?"

Response1: "The Brooklyn Bridge is about 1.1 miles long."

Query2: "When was it opened to the public?"

Response2: "It was opened in December 1883."

Now, it's your turn. Please replace entities in the following conversation using the same process:

Original Conversation:

\textcolor{blue}{\{Input Conversation\}}

\textcolor{blue}{Step 1}: Comprehension Synthesis:

[Identify entities of the conversation]

\textcolor{blue}{Step 2}: Associative Expansion:

[Find suitable replacements for the identified entities]

\textcolor{blue}{Step 3}: Conclusion:

[Reconstruct the conversation with new entities based on the outputs of the last two steps]

\textcolor{blue}{\{Output\}}

\end{prompt}

\begin{prompt}[title={Prompt: Dependency Identifying}, label=prompt:ent]
Task Overview:

Your task is to analyze a given conversation and identify the turns that are necessary for understanding the current search intent. Each turn includes one query and one response. Follow this structured approach:

Example to Illustrate the Process (\textcolor{blue}{Demonstration}):

Original Conversation Context:

Turn1:

Query1: "What are the main attractions in Paris?"

Response1: "The Eiffel Tower and the Louvre are among the top attractions."

Turn2:

Query2: "Is the Louvre open on Sundays?"

Response2: "Yes, it's open from 9 AM to 6 PM."

Current Search Intent (New Query):

Query3: "How can I get tickets to the Louvre?"

\textcolor{blue}{Step 1}: Comprehension Synthesis (Identify key themes and intents)

Output: Theme - Paris attractions; Intent - Acquiring information about attractions and related logistics.

\textcolor{blue}{Step 2}: Associative Expansion (Evaluate the importance of each turn in relation to the current intent)

Output: Turn1: General information about attractions; not directly relevant to ticket acquisition. Turn2: Specific information about the Louvre; more relevant to planning a visit, potentially linked to ticketing information.

\textcolor{blue}{Step 3}: Conclusion (Select turns crucial for the current intent)

Necessary Turns:

Turn2.

Now, it's your turn. Please identify the necessary turns in the following conversation using the same process:

Original Conversation:

\textcolor{blue}{\{Input Conversation\}}

\textcolor{blue}{Step 1}: Comprehension Synthesis:

[Identify key themes and intents of the conversation]

\textcolor{blue}{Step 2}: Associative Expansion:

[Evaluate the importance of each turn in relation to the current intent]

\textcolor{blue}{Step 3}: Conclusion:

[Select turns crucial for the current intent based on the outputs of the last two steps]

\textcolor{blue}{\{Output\}}

\end{prompt}

\begin{prompt}[title={Prompt: Noisy Turn}, label=prompt:noi]
Task Overview:

Your task is to introduce a noisy turn (one query and one response) into an existing conversation. This turn should be relevant to the main background of the original conversation but introduce a new, slightly divergent element. Use the following structured approach:

Example to Illustrate the Process (\textcolor{blue}{Demonstration}):

Original Conversation:

Query1: "Can you tell me about the history of the Sydney Opera House?"

Response1: "Certainly, it was designed by Jørn Utzon and opened in 1973."

Query2: "Is it true that Utzon faced challenges during its construction?"

Response2: "Yes, there were significant design and financial challenges that led to his resignation."

\textcolor{blue}{Step 1}: Comprehension Synthesis (Identify key themes and intents)

Output: Theme - Sydney Opera House's history; Intent - Learning about design, construction challenges, and historical events.

\textcolor{blue}{Step 2}: Associative Expansion (Generate a related but distinct element)

Output: Exploring Utzon's architectural style or other famous works.

\textcolor{blue}{Step 3}: Conclusion (Introduce the new turn)

Noisy Turn:

Query: "Apart from the Sydney Opera House, did Utzon design other notable buildings?"

Response: "Yes, he also designed the Bagsværd Church in Denmark, known for its unique roof structure."

Now, it's your turn. Please introduce a noisy turn into the following conversation using the same process:

Original Conversation:

\textcolor{blue}{\{Input Conversation\}}

\textcolor{blue}{Step 1}: Comprehension Synthesis:

[Identify key themes and intents of the conversation]

\textcolor{blue}{Step 2}: Associative Expansion:

[Generate a related but distinct element of existing ones]

\textcolor{blue}{Step 3}: Conclusion:

[Generate a noisy turn based on the outputs of the last two steps]

\textcolor{blue}{\{Output\}}

\end{prompt}

\begin{prompt}[title={Prompt: Paraphrasing}, label=prompt:para]
Task Overview:

Your task is to paraphrase the provided conversation while preserving the original intent and meaning. Each turn in the conversation, including queries and responses, should be paraphrased thoughtfully.

Example to Illustrate the Process (\textcolor{blue}{Demonstration}):

Original Conversation:

Query1: "What time does the train leave?"

Response1: "The train leaves at 6 PM."

Query2: "Do I need to buy a ticket in advance?"

Response2: "Yes, you need to purchase your ticket early."

Query3: "How early should I arrive at the station?"

\textcolor{blue}{Step 1}: Comprehension Synthesis (Identify key themes and intents)

Output: Theme - Travel logistics; Intent - Acquiring information about train schedules, ticketing, and station arrival time.

\textcolor{blue}{Step 2}: Associative Expansion (Generate alternative expressions based on existing ones)

Output: Train schedule -> Queries about departure times
Ticketing -> Questions about ticket purchase requirements

\textcolor{blue}{Step 3}: Conclusion (Paraphrase the conversation based on outputs of last two steps)

Paraphrased Conversation:

Query1: "What hour is the train scheduled to depart?"

Response1: "The train's departure is set for 18:00."

Query2: "Should I purchase a ticket beforehand?"

Response2: "It‘s recommended to get ticket in advance."

Query3: "What‘s the suggested arrival time at station?"

Now, it's your turn. Please paraphrase the following conversation using the same process:

Original Conversation:

\textcolor{blue}{\{Input Conversation\}}

\textcolor{blue}{Step 1}: Comprehension Synthesis:

[Identify key themes and intents of the conversation]

\textcolor{blue}{Step 2}: Associative Expansion:

[Generate alternative expressions based on existing ones]

\textcolor{blue}{Step 3}: Conclusion:

[Paraphrase the conversation based on outputs of the last two steps]

\textcolor{blue}{\{Output\}}

\end{prompt}

\begin{prompt}[title={Prompt: Intent Shifting}, label=prompt:int]
Task Overview:

Your task is to modify the current conversation by shifting its search intent. The new conversation should retain similar expressions to the original but embody a distinctly different intent. Follow this structured approach:

Example to Illustrate the Process (\textcolor{blue}{Demonstration}):

Query1: "Can you recommend some good Italian restaurants in New York City?"

Response1: "Sure, one popular option is L'Artusi in the West Village."

Query2: "Do they offer vegetarian dishes?"

Response2: "Yes, they have a variety of vegetarian options."

\textcolor{blue}{Step 1}: Comprehension Synthesis (Identify key themes and intents)

Output: Theme - Italian restaurants; Intent - Seeking recommendations in New York City.

\textcolor{blue}{Step 2}: Associative Expansion (Choose a distinctly different intent)

Output: New Intent - Inquiring about Italian cooking classes in New York City.

\textcolor{blue}{Step 3}: Conclusion (Reconstruct the conversation with the new intent)

Intent-Shifted Conversation:

Query1: "Can you suggest some places to learn Italian cooking in New York City?"

Response1: "Certainly, one well-known place is the Culinary Institute in Lower Manhattan."

Query2: "Do they offer classes for beginners?"

Response2: "Yes, they have a variety of courses for beginners."

Now, it's your turn. Please shift the intent of the following conversation using the same process:

Original Conversation:

\textcolor{blue}{\{Input Conversation\}}

\textcolor{blue}{Step 1}: Comprehension Synthesis:

[Identify key themes and intents of the conversation]

\textcolor{blue}{Step 2}: Associative Expansion:

[Shift the intent based on existing ones]

\textcolor{blue}{Step 3}: Conclusion:

[Shift the conversation's intent based on the outputs of the last two steps]

\textcolor{blue}{\{Output\}}

\end{prompt}

\subsection{Other Prompts}
\label{subsec:other_prompt}

\begin{prompt}[title={Prompt: Simple QA to Caluculate Perplexity}, label=prompt:ppl]
Task Overview:

I'm going to provide you with a conversation context and a current query. Your task is to answer the current query based on the information of the context:

Example to Illustrate the Process (\textcolor{blue}{Demonstration}):

Conversation Context:

Query1: Can you recommend an Italian restaurant for me in New York City? 

Response1: Giovanni's Veggie Delight is a popular Italian restaurant in NYC. 

Query2: What's the weather like in New York today?  

Response2: It's currently sunny and warm in New York.   

Current Query:

Great, does Giovanni's have outdoor seating?

Response:

Yes, they have a beautiful patio area.  

Now, it's your turn. Please answer the following conversation:

Conversation Context:

\textcolor{blue}{\{Input Conversation Context\}}

Current Query:

\textcolor{blue}{\{Input Current Query\}}

Response:

\textcolor{blue}{\{Output\}}

\end{prompt}

\begin{prompt}[title={Prompt: Naive Paraphrasing}, label=prompt:naive_para]
Task Overview:

I'm going to provide you with a conversation. Your task is to paraphrase the conversation while keeping the original intent and meaning intact. Each turn in the conversation, including a query and a response, should have a corresponding paraphrased version. Here's an example to illustrate:

Example to Illustrate the Process (\textcolor{blue}{Demonstration}):

Original Conversation:

Query1: "What time does the train leave?"

Response1: "The train leaves at 6 PM."

Query2: "Do I need to buy a ticket in advance?"

Response2: "Yes, you need to purchase your ticket early."

Query3: "How early should I arrive at the station?"

Paraphrased Conversation:

Query1: "What hour is the train scheduled to depart?"

Response1: "The train's departure is set for 18:00."

Query2: "Should I purchase a ticket beforehand?"

Response2: "It‘s recommended to get ticket in advance."

Query3: "What‘s the suggested arrival time at station?"

Now, it's your turn. Please paraphrase the following conversation:

Original Conversation:

\textcolor{blue}{\{Input Conversation\}}

Paraphrased Conversation:

\textcolor{blue}{\{Output\}}

\end{prompt}

\section{Details of Calculating Difficulty}
\label{sec:topic_count}
To estimate a conversation's complexity, we use $\text{Diff}(C) = |T_\text{h}|+\left(|\text{Topic}(C)|*\overline{\text{PPL}(C)}\right) $.
This equation is comprised of three components:
(1) The number of the historical turns $T_\text{h}$.
Longer conversations often contain richer information~\cite{sigir22_COTED}.
(2) The number of topics.
Each new topic introduces potential contextual shifts.
We apply a topic model to count $C$'s topics (more details are in Appendix~\ref{sec:topic_count}).
The topic model we used was pre-trained on Wikipedia (\url{https://huggingface.co/MaartenGr/BERTopic_Wikipedia}).
We illustrate the process of counting topics for a conversation $C$ in Alg.~\ref{algorithm:topic_count}.
Intuitively, we assume the first turn of $C$ has one topic and each turn can only add at most one topic to its previous turn.
To ensure we only count new topics, we only add a topic if our topic model is more confident of identifying this new topic than its last identified topic.
(3) The average perplexity of $C$.
Perplexity is a measure to quantify how well an LM predicts a sample.
We prompt an LLM (Appendix~\ref{subsec:other_prompt}) to predict the response based on the context and compute the average perplexity of all turns. 
A higher $\overline{\text{PPL}(C)}$ indicates that the conversation contains a more challenging language.

The sentence-transformer model we use to calculate the similarity between augmented samples is all-MiniLM-L6-v2 (\url{https://huggingface.co/sentence-transformers/all-MiniLM-L6-v2}).

\begin{algorithm}
\caption{Counting Topics with Confidence}
\begin{algorithmic}
\REQUIRE a conversation $\{t_1, t_2, \ldots, t_n\}$, a topic model $f(\cdot)$

\STATE \textbf{Initialize} $topicCounts$ as an empty list
\STATE \textbf{Initialize} $topics$ as an empty list

\FOR{$i$ in $n$}
    \STATE $P \gets$ $f(\{t_1, \ldots, t_i\})$
    \STATE $P \gets$ $P \setminus topics$
    \STATE $P \gets$ SORT($P$, DESCENDING)
    \IF{$i == 1$}
        \STATE APPEND $1$ TO $topicCounts$
        \STATE APPEND ARGMAX($P$) TO $topics$
        \STATE $confidence \gets$ $P[0]-P[1]$
    \ELSE
        \STATE $confidence' \gets P[0]-P[1]$
        \IF{$confidence' \geq confidence$}
            \STATE APPEND $topicCounts[i-1]+1$ TO $topicCounts$
            \STATE APPEND ARGMAX($P$) TO $topics$
            \STATE $confidence \gets confidence'$
        \ELSE
            \STATE APPEND $topicCounts[i-1]$ TO $topicCounts$
        \ENDIF
    \ENDIF
\ENDFOR
\RETURN $topicCounts$
\end{algorithmic}
\label{algorithm:topic_count}
\end{algorithm}

\section{Examples of Generated Conversations}

\begin{figure*}[!t]
	\centering
	\includegraphics[width=0.95\textwidth]{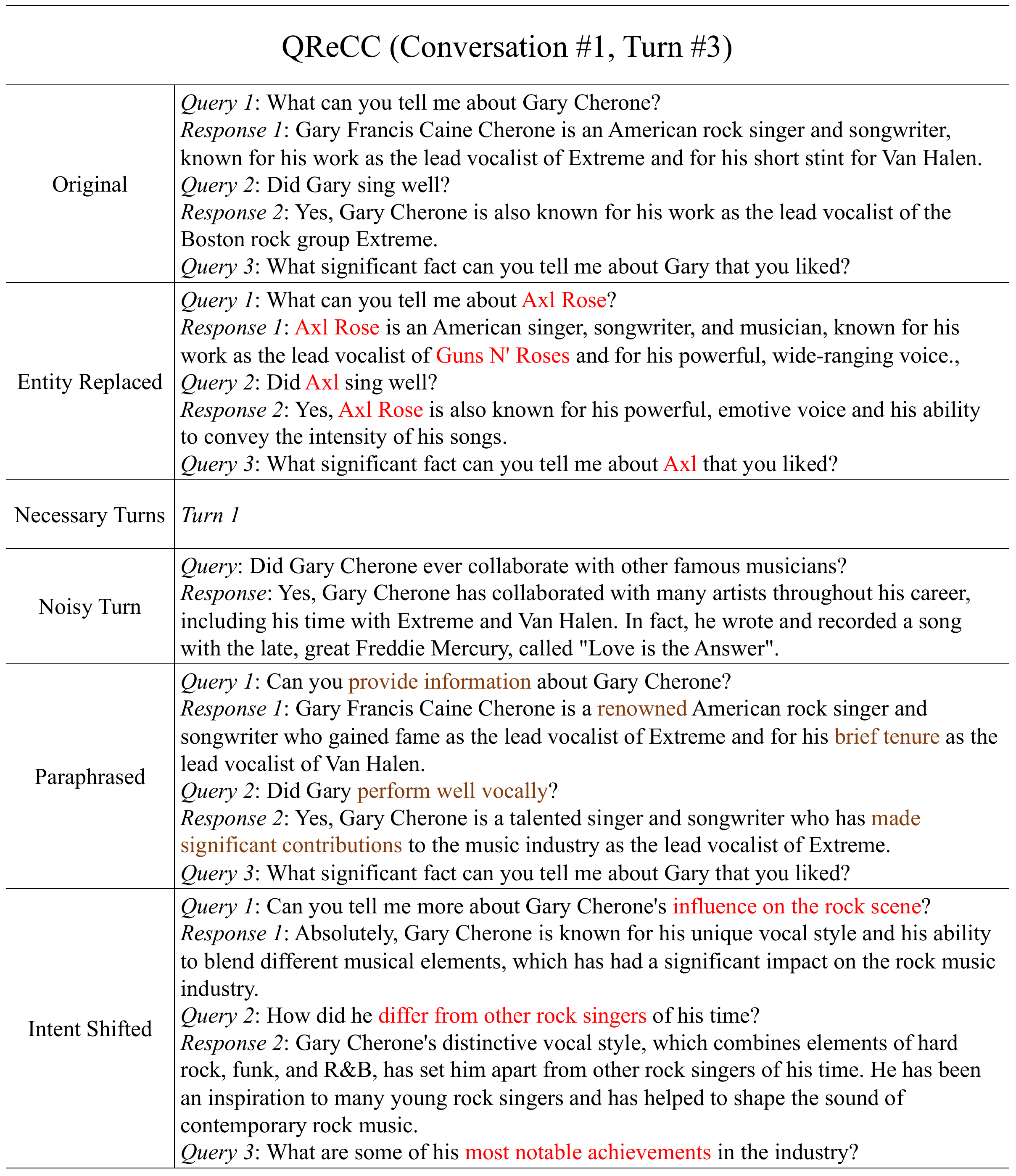}
	\caption{An example to show the generated data of the LLM for a turn in QReCC. }
	\vspace{-2ex}
	\label{fig:qrecc_example}
\end{figure*}

\begin{figure*}[!t]
	\centering
	\includegraphics[width=0.95\textwidth]{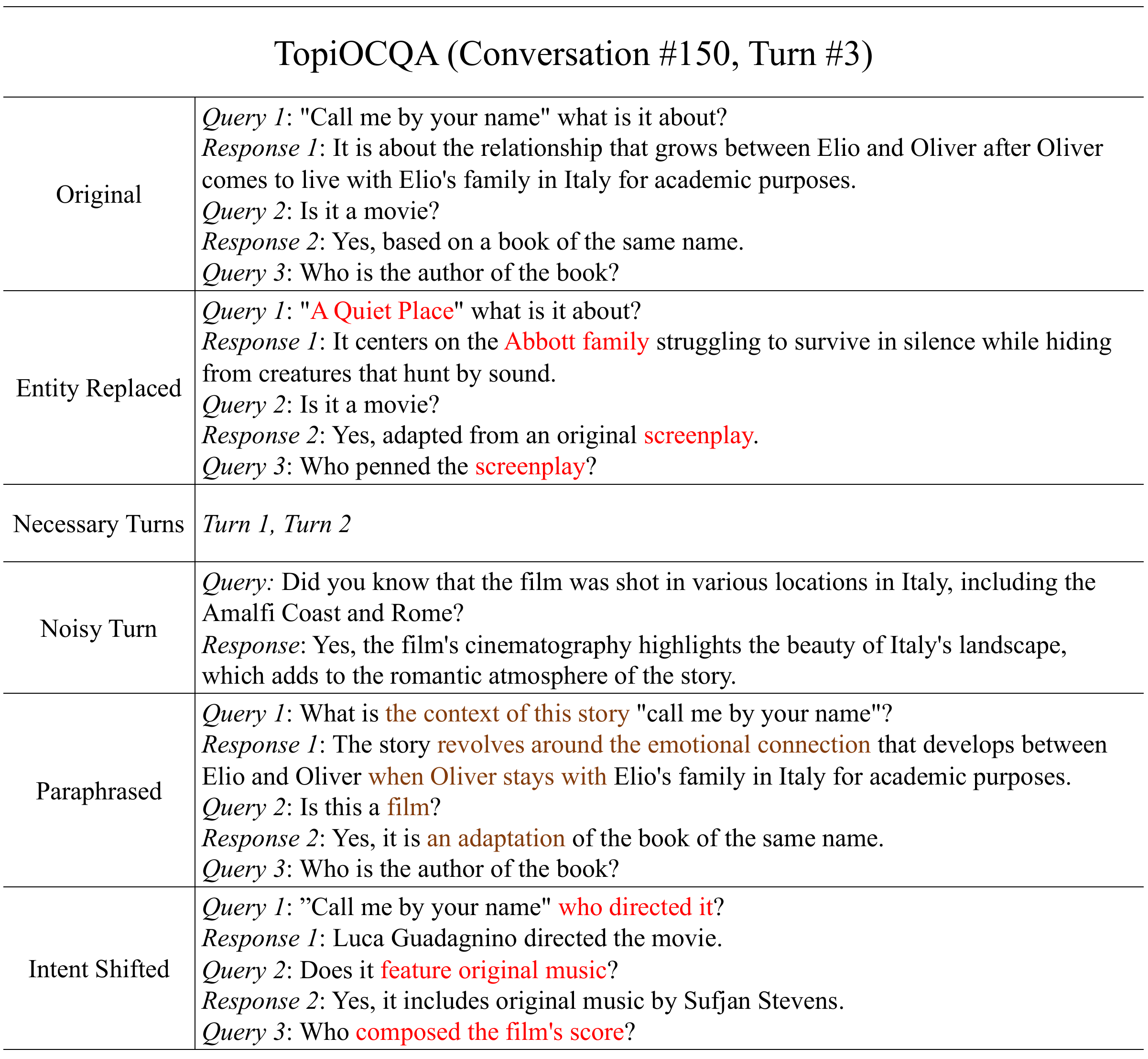}
	\caption{An example to show the generated data of the LLM for a turn in TopiOCQA. }
	\vspace{-2ex}
	\label{fig:topiocqa_example}
\end{figure*}

In this section, we present two examples of the full generated data of a turn by the LLM in Figure~\ref{fig:qrecc_example} and Figure~\ref{fig:topiocqa_example}.
We only show the data generated by the LLM and the example contexts augmented by rule-based strategies (token masking, turn masking, and reordering based on the dependency graph generated by LLM) can be found in Figure~\ref{fig:data_aug}.

\end{document}